\documentclass{article}
\usepackage{spconf,amsmath,graphicx}
\title{Test your samples jointly:  Pseudo-reference for image quality evaluation}

\name{Marcelin Tworski $\dagger$ $\ddagger$, Stéphane Lathuilière $\dagger$}
\address{$\dagger$  LTCI, Télécom-Paris, Institut Polytechnique de Paris; $\ddagger$DXOMARK}
\usepackage{times}
\usepackage{epsfig}
\usepackage{graphicx}
\usepackage{amsmath}
\usepackage{math}
\usepackage{amssymb}
\usepackage{subcaption}
\usepackage[table]{xcolor}
\usepackage{booktabs}
\usepackage{multirow}
\usepackage{color}
\usepackage{cases}
\usepackage{placeins}
\usepackage{pifont}
\usepackage[shortlabels]{enumitem}
\newcommand{\marcelin}[1]{\textcolor{red}{Marcelin: #1}}

\newcommand{\etal}{\textit{et al}. }

\newcommand{\eg}{\textit{e}.\textit{g}. }

\DeclareMathOperator{\GAP}{GAP}

\DeclareMathOperator{\Conv}{Conv}

\DeclareMathOperator{\Softmax}{Softmax}
\DeclareMathOperator{\SSIM}{SSIM}

\begin{document}
\ninept
\maketitle
%
%\begin{small}
%\setlength{\abovedisplayskip}{4pt}
%\setlength{\belowdisplayskip}{2pt}
%\ninept
\begin{abstract}
In this paper, we address the well-known image quality assessment problem but in contrast from existing approaches that predict image quality independently for every images, we propose to jointly model different images depicting the same content to improve the precision of quality estimation. This proposal is motivated by the idea that multiple distorted images can provide information to disambiguate image features related to content and quality. To this aim, we combine the feature representations from the different images to estimate a pseudo-reference that we use to enhance score prediction. Our experiments show that at test-time, our method successfully combines the features from multiple images depicting the same new content, improving estimation quality.
\end{abstract}
\begin{keywords}
Image quality assessment, Joint-evaluation
\end{keywords}

\section{Introduction}

Image quality assessment (IQA) can be addressed in two different settings: no-reference IQA, which consists in estimating the quality of an image without additional information, and full-reference IQA, where we assume that we have at our disposal a high-quality or pristine image that is used to predict the quality of a degraded image.
In this paper, we explore a variant of no-reference IQA where we assume that at test time the goal is to estimate the quality score of different images depicting the same content.
In this setting, we can take advantage of the multiple distorted images by modeling the variability over the different test samples. This allows us to provide content context to the evaluated samples as would a reference, without requiring a reference for the scene. %At test-time, pseudo reference feature maps are computed on the fly and provide good adaption on an unseen scene. This is a no-reference setting where we assume to have multiple inputs at test-time of the same content.

This new setting is motivated by several use cases. It is especially relevant to the case of image quality assessment for camera evaluation. In this task, different cameras are usually compared on the same content \cite{tworski2020dr2s} \cite{carco}. The reference image is not available when it comes to an evaluation in in-the-wild conditions on natural content.  Another example is the case of image enhancement where the reference image is unknown and more accurate quality evaluation algorithms could lead to better enhancement. %Our proposed setting could improve the prediction precision and in consequence algorithm comparisons.

%In consequence, we propose a novel image quality evaluation setting that takes advantage of these novel assumptions. 
To address this new setting, we introduce a new network architecture and its corresponding training strategy. Our architecture allows information exchange across samples of the input batch. More precisely, we train our network to compute a pseudo-reference that describes the evaluated scene. At test time, our method, Pseudo-Reference for Image Quality (PRIQ), is given registered samples of a new scene. The pseudo-reference is predicted by a sub-network that combines features from the different test samples. We perform extensive ablations experiments and compare the performances of the proposed method with state-of-the-art approaches on three different datasets.
%\marcelin{TO REM ?This strategy has the advantage of not requiring access to the source dataset when adapting to the target domain, to require no training, supervised or unsupervised, on the target domain, and no annotations from the target domain.} In short our contributions are \marcelin{Didn't I just said it ?}

\section{Related Work}

%\textbf{no-reference IQA.} 

While  earlier no-reference methods used handcrafted features such as Natural Scene Statistics \cite{brisque, ilniqe}, or handbook of features \cite{cornia}, the best-performing methods in image quality assessment are nowadays learning-based. Even though standard vision architectures provide solid baselines, several domain-specific methods have been proposed:
Bosse \etal proposes to equip their convolutional neural network with a patch weighting estimator. Zeng \etal \cite{PQR} use annotations as a Gaussian distribution around the ground truth instead of a single value score. The  DBCNN architecture \cite{zhang2018blind} proposed by Zhang \etal is composed of two different convolutional models: the first one trained on classification of synthetic distortions while the second one is a pre-trained  VGG \cite{vgg} on ImageNet \cite{deng2009imagenet}, representing perceptual features of natural images. Su \etal \cite{hyperiqa} proposed to disambiguate content features and quality features with the help of a content understanding hypernetwork. Techniques from the computer vision literature have been applied to the image quality assessment task: Meta-learning techniques have been applied by Zhu \etal \cite{metaiqa} in order to improve the performances to unknown distortions. CONTRIQUE \cite{contrique}, proposed by by Madhusudana \etal apply contrastive learning techniques, training their network to produce relevant features for image quality on a large unlabeled dataset. Finally, with the rise of transformers for vision tasks, Ke \etal \cite{musiq} and Golestaneh \etal \cite{tres} proposed novel architectures to tackle the no-reference image quality problem.
Despite remarkable progresses due to more advanced architectures and training procedures, all these methods treat the test samples independently while jointly modeling samples can lead to better predictions.
%\steph{repetitive sentences: XX etal ....Change formulatio}
%\textbf{Use of pseudo-reference in IQA.} 

In no-reference IQA literature, very few works employ a pseudo-reference to devise a formulation closer to full-reference IQA methods. In particular, Lin \etal \cite{halluiqa}  introduce an image generator that estimates an "hallucinated-reference" from a distorted image. In the same vein, Zheng \etal \cite{ckdn} learn a network that projects the reference and any distorted image to the same feature representation. At test time, another distorted image is provided in order to estimate the pseudo-reference features used to evaluate the target image. Although these two methods are based on pseudo-reference estimation, our approach fundamentally differs from these works since we do not need reference images at training time. 

\iffalse \marcelin{Ne pas mettre ?  peut etre phrase en intro : Alors que cs sujets on été abordés en classification,... testtime DA:
- evaluated only for classification problem while we deal with regression
- recognition task while we estimate quality that is a lower level signal processing task
- they require expensive training procedure at test time while our approach is take advantage of its multiple observation without training and an very limited computational overhead }
\fi

\section{Method}
\label{sec:Method}

\begin{figure*}[t]\centering
\includegraphics[width=\textwidth]{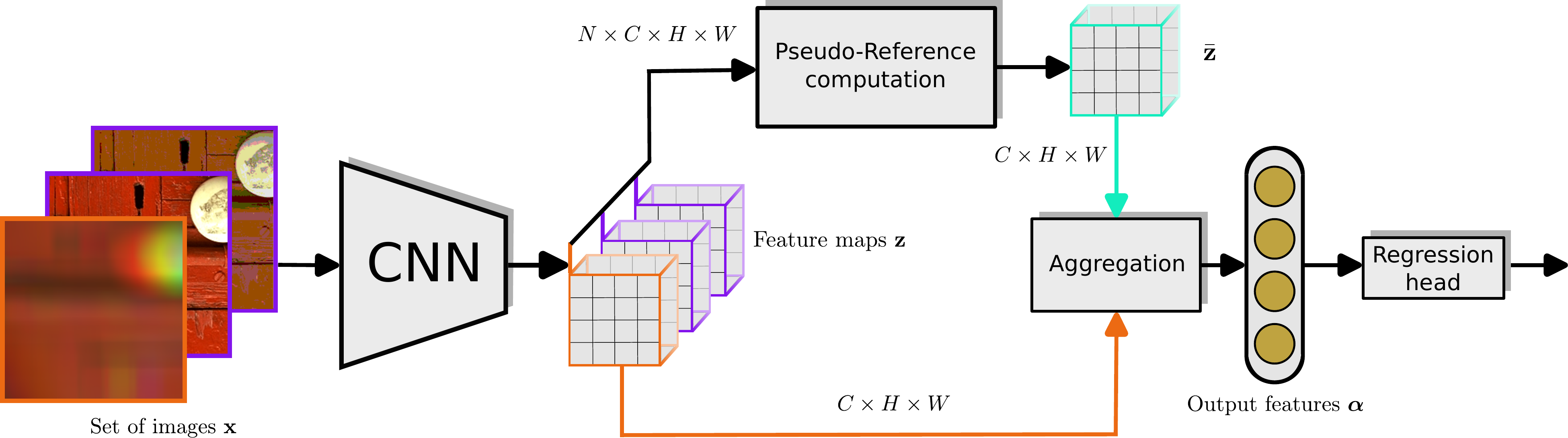}
\caption{Illustration of the proposed model. The evaluation pipeline of the evaluated image is highlighted in red. In practice this procedure is applied in parallel for all the images in the batch. After a transformation of the set of images into feature maps, all these feature maps are used to compute the pseudo-reference, which is then aggregated with the evaluated image's feature maps to produce the output features used to predict quality. \label{fig:pipeline}}\end{figure*}
%\marcelin{We assume that we have at our disposal $D$ source domains $\{\mathcal{S}_1, ...,\mathcal{S}_D\}$. Each source domain $\mathcal{S}_d$ contains $N_d$ images $\mathcal{X}_d=\{\xmat_d^n\in\mathbb{R}^{H\times W\times 3}, 1\leq\! n\leq\! N_d\}$ with their corresponding quality scores $\mathcal{Y}_d=\{y_d^n, 1\leq\!n\!\leq N_d\}$.Note that, we consider that images of the same domain depict the same content. At training time we only assume to have access to the source domains. At test time, we have at our disposal test images without annotations.}

%\steph{should we say scene or content. I think I prefer content/ D different distortions of the same content}
%\marcelin{We should not say the word distortions. It gives the impression our method need distortions labels}

In this work, we assume that we have at our disposal $D$ different scenes $\{\mathcal{S}_1, ...,\mathcal{S}_D\}$ with each scene containing $N_d$ samples of different qualities depicting the same content. Our goal is to estimate image quality scores on novel scenes. While existing methods individually test each sample, our method jointly predicts the score of several samples. 
We design our approach to enable a joint prediction at test time: while each image will be mapped to a different quality score, its prediction depends on all images in the set. The underlying idea is that, since the samples correspond to different unknown distortions, they can provide contextual information and allow the network to differentiate image features related to content from features related to quality.

To achieve this objective, we introduce the architecture illustrated in Fig.  \ref{fig:pipeline} where  predictions depend on other images in the input set. More specifically, after projecting the images into feature maps, these images are combined into a single pseudo-reference, of the same shape as the feature maps of one image. For each image, these pseudo-reference feature maps are aggregated with each image of the set with using methods inspired by the full-reference literature\cite{gao2017deepsim, carco,varga2020combined}. These features are then fed into a single-layer regression head which predicts the image quality scores.

\vspace{-0.1cm}
\subsection{Pseudo-Reference computation}

Let us assume a convolutional backbone network and a set of $N$ images ${\xmat^d_i}$, $i \in [1;N]$, $d \in [1;D]$ from the $d^{th}$ scene. At training time, this set is randomly sampled from the $N_d$ training images of the corresponding scene. At test time, the user can use all the images available. At an intermediary layer of the convolutional network, this set is represented by a tensor $\zmat = (z_{ickl})$, of dimension $N \times C \times H \times W$, where $C$ denotes the number of channels per image at this specific network layer, and $H$ and $W$ are the height and width of the tensor. We aim to compute a tensor $\bar{\zmat}$, of dimension $C \times H \times W$, acting as a pseudo-reference for this set of images.

We propose to estimate the pseudo reference with a weighted mean from the feature maps of each image in the set. In this aim, we compute the weights $\wmat$ measuring the relevance the network should give to each image. More precisely, we perform this computation for every image and location to obtain a pseudo-reference that fully benefits from all the images in the set. Therefore, $\overline{\zmat}$ is given by:
\begin{equation}
\label{theeq}
\overline{\zmat} = \sum_{i=1}^N \wmat_{i} \odot \zmat_{i},
\end{equation}
with $\odot$ being the element-wise multiplication in the $H \times W$ dimensions. The weights $\wmat$ are predicted by a sub-network with the feature maps $\zmat$ as inputs through an attention mechanism. More precisely, they are computed with a one-by-one convolutional layer with one kernel in order to output one channel. The weights need to sum to one over the set dimension to implement the weighted mean operation. Therefore, we follow common practices in attention models and employ the softmax activation:
\begin{equation}
    \wmat = \Softmax(\Conv(\zmat)))
\end{equation}
%with the softmax operator being applied in the set dimension,
Note that this pseudo-reference is computed once with every image in the set, and the same pseudo-reference is used to evaluate every image in the set.

\subsection{Aggregation}

After describing the pseudo-reference computation,  we can now detail the aggregation scheme that we employ to compare every input image to the pseudo-reference.
The aggregation layer receives as inputs the features maps $\zmat_i$ for an image $i$ in the set and the pseudo-reference feature maps $\bar{\zmat}$ and outputs a feature vector for each input image people. Inspired by the full-reference literature \cite{gao2017deepsim,carco,varga2020combined}, we chose to apply the channel-wise Structural Similarity Index Measure (SSIM).

To compare the two tensors, we treat separately each channel and consider the two feature maps as $C$ samples of two random variables. Assuming that $\zmat_{ic}$ and $\bar{\zmat}_{c}$ $\in \mathbb{R}^{H \times W}$  respectively, denote the values in $\zmat_i$ and $\bar{\zmat}$, the output vector $\alphavect=\big(\alpha_c\big)_{1\leq c \leq C}$ for the i-th sample of the SSIM aggregation layer is given by:
\begin{equation}
  \alpha_c = \SSIM(\zmat_{ic}, \bar{\zmat}_{c})
\end{equation}
%For each image in the batch, the output features of the aggregation layer is given by:
%\begin{equation}
%    \betavect = \alphavect \cdot \left(\GAP(z_ic)_{c \in [1;C]}\right)
%\end{equation}
%with $\cdot$ the concatenation operator. The length of $\betavect$ is $2C$
%Indeed, preliminary experiments showed that adding features directly from the image we want to have a prediction helps in a low-data setting where the test batch size is small.
\subsection{Feature pyramid and prediction}
Selecting the right scale for comparing the images with  the pseudo-reference is not straightforward. We propose to employ a feature pyramid strategy \cite{lin2017feature, musiq, carco} which provides a set of features that are used to evaluate image quality at different scales. The pseudo-reference computation and the aggregation are applied to the output of 5 different intermediate layers of the backbone network at different resolutions. The extracted features at each resolution are then concatenated and fed to a regression head for the image quality prediction.

\section{Experiments}
\subsection{Evaluation Protocol}

\textbf{Datasets.} To evaluate the proposed method, we perform experiments on three datasets:
%\begin{itemize}[noitemsep,topsep=-2pt,parsep=0pt,partopsep=0pt,labelwidth=1cm,align=left,itemindent=0cm,leftmargin=*]
\begin{itemize}
\item TID2013~\cite{ponomarenko2015image} consists of 25 different contents and 24 distortions with 5 levels each, for a total of 3000 images 
\item KADID10k~\cite{lin2019kadid} consists of 81 different contents and 25 distortions with 5 levels each, for a total of 10125 images
\item CARCO dataset~\cite{carco} presents distortions from numerous camera devices. It differs from other in-the-wild datasets which do not present the same content with varying qualities \cite{hosu2020koniq, 7327186} or lack of region selection and registration \cite{fang2020perceptual}.
\end{itemize}

The images of these three datasets are registered. Registration is exact by design for TID2013 and KADID10k while the input images were approximately registered by a prepossessing algorithm for CARCO.

\noindent\textbf{Metrics. }Regarding evaluation metrics, we follow previous works \cite{brisque, niqe} and use the Linear Correlation Coefficient (LCC) between the annotations and predictions. Additionally, we report the Spearman Rank-Order Correlation Coefficient (\emph{SROCC}) defined as the linear correlation coefficient of the ranks of predictions and annotations. We report the median of these metrics across the different runs as in \cite{tres, contrique}.

 \noindent\textbf{Protocol. }In the TID2013 and KADID10k experiments, we select randomly 80\% of the datasets' scenes for training and 20\% for testing. This protocol ensures there is no content overlap between the training and testing sets. We repeat this process five times and we report the median. The split is fixed over multiple experiments, ensuring different experiments are compared with the same split configuration. 
For the experiments on CARCO, the dataset size allows us to effectively test the ten scenes independently as the test scene in a ten-fold cross-validation. This process is also repeated five times for more robust results and we also report the median. 
%\marcelin{maybe to remove and not do itSince our goal is to perform domain adaptation on an unknown new scene, we also report for KADID10K and TID2013 the average of these metrics computed for each scene independently rather than only the computation of the correlation for all the images in the test set.}

% At test-time, we consider that the user wants to evaluate $T$ images. The number of images $T$ corresponds to the set size $N$ at training time. To simulate multiple test experiments, we randomly split our test dataset into multiple sets of size $T$. To ensure fair comparison, in every experiment, we employ the same random sub-sets for all the methods but sample different sub-sets for each run.\steph{correct?} The reported correlation metrics are computed for all the samples of the test dataset jointly. 

At test-time, we simulate different scenarios for a user based on the number of images $T$ available. The number of images $T$ is analogous to the set size $N$ at training time. To simulate these conditions, we randomly split our test dataset into multiple sets of size $T$. To ensure a fair comparison, we employ the same random sets for every method. The reported correlation metrics are computed for all the samples of the test dataset jointly. Note that 
set size $T$ at test-time does not correspond to the size of our test dataset.

\vspace{-0.5cm}
\subsection{Implementation Details.}
\vspace{-0.2cm}

We employ a ResNet-18\cite{he2016deep} backbone, pre-trained on ImageNet \cite{deng2009imagenet}. The chosen five pyramid stages are placed after the initial $7 \times 7$ convolution and after each residual block. We use the Huber loss and train for 60 epochs, with a weight decay of $3 \times 10^{-3}$ every ten epochs. We used a training batch of 30 images, composed of 6 sets of $N =5$ images per batch. %Indeed, on some of the datasets using only one content leads to a lack of diversity and thus an adverse effect on the batch normalization. 
The images are randomly cropped to a $224 \cdot 224$ size and randomly flipped. All the images of a set are augmented in the same manner in order to preserve alignment.

\vspace{-0.5cm}
\subsection{Analysis: Pseudo-Reference Computation.}
\vspace{-0.2cm}

We now evaluate different approaches to compute the pseudo-reference feature maps. We explore different solutions regarding the dimensionality of the weights used in the weighted average used to estimate the pseudo-reference.

\begin{table}[t]
\begin{center}
\tiny

\resizebox{\columnwidth}{!}{ \begin{tabular}{c p{0.01\textwidth} p{0.01\textwidth}c p{0.01\textwidth} c c } 
\toprule
\multirow{2}{*}{\textbf{Module}}&\multicolumn{3}{c}{\textbf{Weights dimension} }& \multirow{2}{*}{\textbf{T}} & \multirow{2}{*}{\textbf{LCC}}& \multirow{2}{*}{\textbf{SROCC}}\\
\\
&$N$&$C$&$H\times W$\\
  \midrule 
 
\textbf{(i)}& \ding{55} &\ding{55}& \ding{55}&\multirow{5}{*}{5} &0.890& 0.850\\
\textbf{(ii)} &\ding{51} &\ding{55}& \ding{55} &&   0.873 &  0.849\\
\textbf{(iii)} &\ding{51} &\ding{51}& \ding{55}&&  \textbf{0.899} &  0.847 \\
\textbf{(iv)}&\ding{51} &\ding{55}& \ding{51}&&  \textbf{0.899} &  \textbf{0.861} \\
\textbf{(v)}&\ding{51} &\ding{51}& \ding{51}& & 0.897&0.856\\

\midrule

\textbf{(i)}&\ding{55} &\ding{55}& \ding{55}&\multirow{5}{*}{20} &0.904 & 0.879\\
\textbf{(ii)}&\ding{51} &\ding{55}& \ding{55} &&  0.893 & 0.872 \\
\textbf{(iii)} &\ding{51} &\ding{51}& \ding{55}&& 0.911& 0.870\\
\textbf{(iv)}& \ding{51} &\ding{55}& \ding{51}&& \textbf{0.918} & \textbf{0.883}\\
\textbf{(v)}&\ding{51} &\ding{51}& \ding{51}& &0.908& 0.879\\
\midrule

\textbf{(i)}&\ding{55} &\ding{55}& \ding{55}&\multirow{5}{*}{100} &0.906 & 0.887\\
\textbf{(ii)} &\ding{51} &\ding{55}& \ding{55} && 0.913&  0.883\\
\textbf{(iii)} &\ding{51} &\ding{51}& \ding{55}&& 0.909& 0.881\\
\textbf{(iv)}& \ding{51} &\ding{55}& \ding{51}&& \textbf{0.926} & \textbf{0.899}\\
\textbf{(v)}& \ding{51} &\ding{51}& \ding{51}&&0.908&0.886\\
 \bottomrule
\end{tabular}}
\caption{Analysis of performances of various modules producing a pseudo-reference. \textit{T} represents the set size at test time. The median correlations over 5 runs are reported.}
\label{tab:ablation}
\end{center}
\vspace{-0.5cm}
\end{table}

In section \ref{sec:Method}, the pseudo-reference is computed following equation \eqref{theeq}. In this analysis, we evaluate this design choice and consider the following variants:

\begin{enumerate}[\bf(i)]
\item A first naive approach is to compute the mean along the set axis: $\bar{\zmat} = \frac{1}{N}\sum_{i=1}^N \zmat_i$.
\item We consider a slightly more complex solution based on a weighted average:
$ \bar{\zmat} = \sum_{i=1}^N \wmat_i \zmat_i$, with $\wmat \in [0,1]^N$.
The weights are computed through a linear layer: $
    \wmat = \Softmax(\Amat(\GAP(\zmat) + \bvect))$, 
with $\Amat$ and $\bvect$ being the parameters of a linear layer with an output size of one, and $\GAP$ designating the \emph{Global Average Pooling} operation, with a Softmax computed along the set axis.
\item  We modify the previous approach to allow different weights for channels of the set images. Here, the output size of the linear layer is $C$, and $\wmat$ is of size $N \times C$.
\item Instead of weighting the channels, we weigh in this module the feature maps' locations. This corresponds to our proposed model as described in Sec. \ref{sec:Method}.
\item It is also possible to weigh the locations and channels simultaneously. Thus, the convolutional layer from \textbf{(iv)} has now $C$ different kernels instead of one and thus outputs feature maps with $C$ channel for each image. In this case $\wmat$ is of size $N \times C \times H \times W$.
\end{enumerate}

The results are reported in Table \ref{tab:ablation}.
We observe that the setting with the weights $\wmat$ in $\mathbb{R}^{N \times H \times W}$ provides the best result for every value of $T$ (\eg 0.013 SROCC increase for 20 samples compared to the second best performing module). Adding the channel dimension to the set of weights generally degrades performances. A simple mean provides good results and is the second or third-best-performing module depending on the set size. 

\begin{figure}[h]\centering
\vspace{-0.5cm}
\includegraphics[width=\columnwidth]{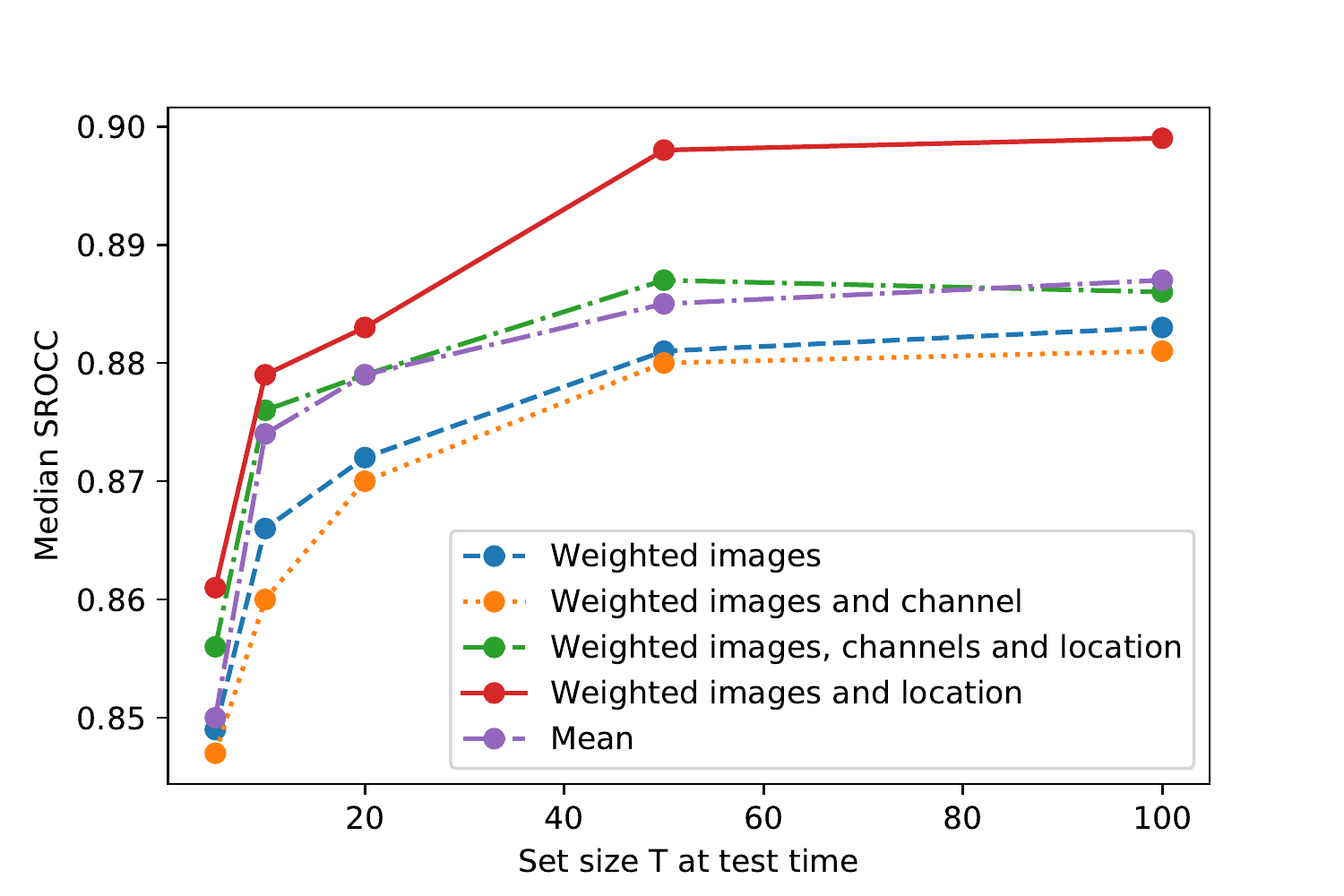} 
\caption{Impact of the set size $T$ at test time for the different pseudo-reference modes.}
\label{tab:thegraph}
\vspace{-0.7cm}
\end{figure}

In Fig.\ref{tab:thegraph}, we provide more values for the set size at test time $T$ to compare these five different modules. These experiments confirm our analysis for intermediary values. Even though our model is trained with a set size of 5, we observe that performances are consistently increasing with the set size at test time. We observe rapid improvements from 2 samples to 10 samples, then a steady rise in performance up to 50 samples, and the results at 100 samples show slight amelioration over the results at 50 samples.

\subsection{Ablation Study}
\vspace{-0.2cm}
We now validate the positive impact of each component of the proposed approach. We consider several variants of our approach. First, we train a vanilla ResNet-18 on our datasets, with no pseudo-reference method. Then, we test our method without the feature's pyramid. Instead, a single pseudo-reference is estimated after the last convolutional block. Finally, we replace the SSIM aggregation of our method with a concatenation of features of the image to be evaluated and the pseudo-reference after an average pooling.

\begin{table}[h]
\begin{center}
\resizebox{\columnwidth}{!}{ \begin{tabular}{ p{0.02\textwidth} p{0.03\textwidth}p{0.02\textwidth}cccccc } 
\toprule
\multirow{2}{*}{\textbf{P-R} }& \multirow{2}{*}{\textbf{SSIM}} & \multirow{2}{*}{\textbf{Pyr}} &\multicolumn{2}{c}{TID2013}& \multicolumn{2}{c}{KADID10k} & 
\multicolumn{2}{c}{CARCO}\\
&&&LCC & SROCC&LCC & SROCC& LCC & SROCC\\
  \midrule 
\ding{55} & \ding{55} & \ding{55} &0.815& 0.782& 0.825 &0.839& 0.883&0.853\\
\ding{51} & \ding{51} & \ding{55} & 0.923 & 0.906 & 0.934 & 0.93&0.903 &0.864\\
\ding{51} & \ding{55} & \ding{51} &0.886&0.873 &0.883 & 0.884 &0.904& 0.873\\
\ding{51} & \ding{51} & \ding{51} & \textbf{0.929}& \textbf{0.911}& \textbf{0.937}&\textbf{0.936} &\textbf{0.918}& \textbf{0.883}\\

 \bottomrule
\end{tabular}}
\caption{Ablation study on the \emph{CARCO} dataset: architectural design. We set $T = 20$ for this comparison. The median correlations over 5 runs are reported.}
\label{tab:ablation2}

\end{center}
\end{table}

For all three dataset we found that our method vastly outperforms a Resnet-18 trained regularly. The SSIM aggregation provides huge improvement over the features' concatenation  on the performances in the case of KADID10k and TID2013, while the improvement is still sizeable on CARCO. Finally, even though the feature pyramid provides only small improvements on the synthetic datasets (TID2013 and KADID10k), the results are still consistently better with the pyramid and its effect on CARCO is important.

\vspace{-0.5cm}
\subsection{Comparison to the State of the Art}
\vspace{-0.2cm}

The comparison to the state-of-the-art is summarized in Table \ref{tab:SOTA}. Our framework is the first approach that can jointly evaluate multiple images jointly. Therefore, our approach is compared with traditional methods that evaluate images independently. We observe that joint evaluation with our method greatly improves predictions according to every metric with $T \geq 5$ for TID2013, and CARCO, whereas CONTRIQUE \cite{contrique} achieves great results on KADID10k and achieves performances roughly equal to our method. Notably, we observe that using a set size larger than 20 for our offers little improvement in the case of synthetic datasets. However, the gain in performance might be interesting on authentically distorted images. Concerning variants explored in Tab. 2, the variant without the pyramid still vastly outperforms other methods on the TID dataset, while only CONTRIQUE is performing similarly on KADID. The variant without the SSIM aggregation is still slightly the best-performing method on TID with the TReS method coming close, while on KADID it is only outperformed by CONTRIQUE. On the CARCO dataset the no-SSIM performs similarly to the best state of the art method while the no-pyramid variant slightly underperforms the 3 best performing methods in terms of SROCC.

We also observe that our method vastly outperforms other approaches on CARCO whose images are authentic camera photographs. This result validates our approach in this particularly relevant setting of camera evaluation.
\begin{table}[h]
\begin{center}
\resizebox{\columnwidth}{!}{ \begin{tabular}{ ccccccc } 
\toprule
&\multicolumn{2}{c}{TID2013}& \multicolumn{2}{c}{KADID10k} & 
\multicolumn{2}{c}{CARCO}\\
\textbf{Method}  & LCC & SROCC & LCC & SROCC & LCC & SROCC\\
  \midrule 
 \emph{BRISQUE} \cite{brisque}  & 0.571&0.626 & 0.567 & 0.528&0.151&0.221\\
 \emph{IL-NIQE} \cite{ilniqe} & 0.648& 0.521& 0.558 & 0.534&0.622&0.499\\
 \midrule
 \emph{ResNet-18} \cite{he2016deep} & 0.815 & 0.782 & 0.825& 0.839 &0.883&0.853\\
 \emph{PQR} \cite{PQR} & 0.798&0.740  & - &  - & - & - \\
\emph{WaDIQaM} \cite{Wadiqam} & 0.855&0.835 & 0.752 & 0.739 & - &- \\
\emph{DBCNN}  \cite{zhang2018blind} & 0.865& 0.816&  0.856&0.851 &0.827&0.746\\
\emph{Meta-IQA} \cite{metaiqa} & 0.868& 0.856& 0.775&0.762&-&-\\
\emph{HyperIQA} \cite{hyperiqa} & 0.858& 0.840& 0.845&0.852 &0.889&0.881\\
\emph{TReS}  \cite{tres} &0.883 & 0.863& 0.858& 0.859&0.904&0.869\\
\emph{CONTRIQUE} \cite{contrique} & 0.857 & 0.843 & \textbf{0.937}& 0.934&0.900& 0.875\\
\midrule
\emph{PRIQ, T=2} &0.857 & 0.841 &0.874 & 0.869 & 0.899 & 0.861\\
\emph{PRIQ, T=5} &0.916 &0.894&0.933&0.934& 0.911 & 0.879 \\
\emph{PRIQ, T=20} &0.929& \textbf{0.911} &\textbf{0.937}& \textbf{0.936}& 0.918 & 0.883\\
\emph{PRIQ, T=100} & \textbf{0.930} &\textbf{0.911}& \textbf{0.937}& 0.935& \textbf{0.926}& \textbf{0.899} \\
\end{tabular}}
\caption{Comparison to the state-of-the-art. The median correlations over 5 runs are reported. Some results are borrowed from \cite{tres} and \cite{contrique}}
\label{tab:SOTA}
\end{center}
\vspace{-0.7cm}
\end{table}
%\subsection{Cross-dataset experiments}

\vspace{-0.5cm}
\section{Conclusion}

We introduced a setting where images are jointly evaluated that effectively uses several different images of the same content in order to provide semantic contextual information. The proposed design estimates a pseudo-reference at feature level, and employs a feature pyramid aggregation. We conducted an ablation experiment to determine the optimal pseudo-reference computation module and another ablation to understand the contribution of each part of our method. We performed extensive evaluations  across several image quality datasets to validate the efficiency of the proposed method, and found that we achieve competitive or better performances than state-of-the-art no-reference image quality methods whether on synthetic or in-the-wild datasets. We encourage IQA researchers to explore this new setting due to the large possibilities for taking advantages of the multiple inputs and due to the possible applications. 
%

%

\bibliographystyle{ieee}
\bibliography{bibliography}

\begin{thebibliography}{10}\itemsep=-1pt

\bibitem{Wadiqam}
Sebastian Bosse, Dominique Maniry, Klaus-Robert M{\"u}ller, Thomas Wiegand, and
  Wojciech Samek.
\newblock Deep neural networks for no-reference and full-reference image
  quality assessment.
\newblock {\em IEEE TIP}, 2017.

\bibitem{deng2009imagenet}
Jia Deng, Wei Dong, Richard Socher, Li-Jia Li, Kai Li, and Li Fei-Fei.
\newblock Image-net: A large-scale hierarchical image database.
\newblock In {\em IEEE CVPR}, 2009.

\bibitem{fang2020perceptual}
Yuming Fang, Hanwei Zhu, Yan Zeng, Kede Ma, and Zhou Wang.
\newblock Perceptual quality assessment of smartphone photography.
\newblock In {\em IEEE CVPR}, 2020.

\bibitem{gao2017deepsim}
Fei Gao, Yi Wang, Panpeng Li, Min Tan, Jun Yu, and Yani Zhu.
\newblock Deepsim: Deep similarity for image quality assessment.
\newblock {\em Elsevier Neurocomputing}, 2017.

\bibitem{7327186}
D. {Ghadiyaram} and A.~C. {Bovik}.
\newblock Massive online crowdsourced study of subjective and objective picture
  quality.
\newblock {\em IEEE TIP}, 2016.

\bibitem{tres}
S~Alireza Golestaneh, Saba Dadsetan, and Kris~M Kitani.
\newblock No-reference image quality assessment via transformers, relative
  ranking, and self-consistency.
\newblock In {\em WACV}, 2022.

\bibitem{he2016deep}
Kaiming He, Xiangyu Zhang, Shaoqing Ren, and Jian Sun.
\newblock Deep residual learning for image recognition.
\newblock In {\em IEEE TPAMI}, 2016.

\bibitem{hosu2020koniq}
Vlad Hosu, Hanhe Lin, Tamas Sziranyi, and Dietmar Saupe.
\newblock Koniq-10k: An ecologically valid database for deep learning of blind
  image quality assessment.
\newblock {\em IEEE TIP}, 2020.

\bibitem{musiq}
Junjie Ke, Qifei Wang, Yilin Wang, Peyman Milanfar, and Feng Yang.
\newblock Musiq: Multi-scale image quality transformer.
\newblock In {\em IEEE CVPR}, 2021.

\bibitem{lin2019kadid}
Hanhe Lin, Vlad Hosu, and Dietmar Saupe.
\newblock Kadid-10k: A large-scale artificially distorted iqa database.
\newblock In {\em QoMEX}. IEEE, 2019.

\bibitem{halluiqa}
Kwan-Yee Lin and Guanxiang Wang.
\newblock Hallucinated-iqa: No-reference image quality assessment via
  adversarial learning.
\newblock In {\em IEEE CVPR}, 2018.

\bibitem{lin2017feature}
Tsung-Yi Lin, Piotr Doll{\'a}r, Ross Girshick, Kaiming He, Bharath Hariharan,
  and Serge Belongie.
\newblock Feature pyramid networks for object detection.
\newblock In {\em IEEE CVPR}, 2017.

\bibitem{contrique}
Pavan~C Madhusudana, Neil Birkbeck, Yilin Wang, Balu Adsumilli, and Alan~C
  Bovik.
\newblock Image quality assessment using contrastive learning.
\newblock {\em TIP}, 2022.

\bibitem{brisque}
Anish Mittal, Anush~Krishna Moorthy, and Alan~Conrad Bovik.
\newblock No-reference image quality assessment in the spatial domain.
\newblock {\em IEEE TIP}, 2012.

\bibitem{niqe}
Anish Mittal, Rajiv Soundararajan, and Alan~C Bovik.
\newblock Making a “completely blind” image quality analyzer.
\newblock {\em IEEE Signal processing letters}, 2012.

\bibitem{ponomarenko2015image}
Nikolay Ponomarenko, Lina Jin, Oleg Ieremeiev, Vladimir Lukin, Karen
  Egiazarian, Jaakko Astola, Benoit Vozel, Kacem Chehdi, Marco Carli, Federica
  Battisti, et~al.
\newblock Image database tid2013: Peculiarities, results and perspectives.
\newblock {\em Elsevier SPIC}, 2015.

\bibitem{vgg}
Karen Simonyan and Andrew Zisserman.
\newblock Very deep convolutional networks for large-scale image recognition.
\newblock In {\em ICLR}, 2015.

\bibitem{hyperiqa}
Shaolin Su, Qingsen Yan, Yu Zhu, Cheng Zhang, Xin Ge, Jinqiu Sun, and Yanning
  Zhang.
\newblock Blindly assess image quality in the wild guided by a self-adaptive
  hyper network.
\newblock In {\em IEEE CVPR}, 2020.

\bibitem{tworski2020dr2s}
Marcelin Tworski, Stéphane Lathuilière, Salim Belkarfa, Attilio Fiandrotti,
  and Marco Cagnazzo.
\newblock Dr2s : Deep regression with region selection for camera quality
  evaluation.
\newblock In {\em ICPR}, 2020.

\bibitem{carco}
Marcelin Tworski, Benoit Pochon, and St{\'e}phane Lathuili{\`e}re.
\newblock Camera quality assessment in real-world conditions.
\newblock {\em Available at SSRN 4166549}, 2022.

\bibitem{varga2020combined}
Domonkos Varga.
\newblock A combined full-reference image quality assessment method based on
  convolutional activation maps.
\newblock {\em Algorithms}, 2020.

\bibitem{cornia}
Peng Ye, Jayant Kumar, Le Kang, and David Doermann.
\newblock Unsupervised feature learning framework for no-reference image
  quality assessment.
\newblock In {\em IEEE CVPR}. IEEE, 2012.

\bibitem{PQR}
Hui Zeng, Lei Zhang, and Alan~C Bovik.
\newblock Blind image quality assessment with a probabilistic quality
  representation.
\newblock In {\em IEEE ICIP}. IEEE, 2018.

\bibitem{ilniqe}
Lin Zhang, Lei Zhang, and Alan~C Bovik.
\newblock A feature-enriched completely blind image quality evaluator.
\newblock {\em IEEE TIP}, 2015.

\bibitem{zhang2018blind}
Weixia Zhang, Kede Ma, Jia Yan, Dexiang Deng, and Zhou Wang.
\newblock Blind image quality assessment using a deep bilinear convolutional
  neural network.
\newblock {\em IEEE TCSVT}, 2018.

\bibitem{ckdn}
Heliang Zheng, Huan Yang, Jianlong Fu, Zheng-Jun Zha, and Jiebo Luo.
\newblock Learning conditional knowledge distillation for degraded-reference
  image quality assessment.
\newblock In {\em IEEE ICCV}, 2021.

\bibitem{metaiqa}
Hancheng Zhu, Leida Li, Jinjian Wu, Weisheng Dong, and Guangming Shi.
\newblock Metaiqa: Deep meta-learning for no-reference image quality
  assessment.
\newblock In {\em CVPR}, 2020.

\end{thebibliography}
%\bibliography{strings,refs}
%\end{small}

\end{document}